\documentclass[conference]{IEEEtran}
\IEEEoverridecommandlockouts
\usepackage{cite}
\usepackage{amsmath,amssymb,amsfonts}
\usepackage{algorithmic}
\usepackage{graphicx}
\usepackage{textcomp}
\usepackage{xcolor}
\usepackage{algorithm}
\usepackage{algorithmic}

\usepackage{algorithm}
\usepackage{algorithmic}
\usepackage{amsmath} 
\usepackage{amssymb} 
\usepackage{graphicx}
\usepackage{booktabs}
\usepackage{graphicx}
\usepackage{amssymb}
\usepackage{booktabs}
\usepackage{dsfont}
\usepackage{amsfonts,amssymb}
\usepackage{graphics}
\usepackage{tikz}
\usepackage{multirow, graphicx}
\usepackage{colortbl}
\usepackage{bbding}
\usepackage{bm}
\usepackage{algorithm}
\usepackage{algorithmic}
\usepackage{amsmath}
\usepackage{multirow, graphicx}
\usepackage{colortbl}
\usepackage{bbding}
\usepackage{bm}

\usepackage{amsmath}

\usepackage{color,xcolor, colortbl}
\definecolor{Gray2}{gray}{0.95}
\definecolor{blue1}{RGB}{68,114,196}
\definecolor{blue2}{RGB}{198,212,238}
\definecolor{orange1}{RGB}{252,163,19}
\definecolor{orange2}{RGB}{252,219,180}
\definecolor{red1}{RGB}{192,0,0}
\definecolor{red2}{RGB}{237,179,178}
\usepackage{adjustbox}
\usepackage{pifont} 
\usepackage{overpic}
\usepackage{subfloat}
\usepackage{enumitem}
\usepackage{makecell}
\usepackage{array}
\usepackage{caption}
\usepackage{algorithm}
\usepackage{algorithmic}
\usepackage{textcomp}

\usepackage{amsmath}

\usepackage[switch]{lineno}
\def\BibTeX{{\rm B\kern-.05em{\sc i\kern-.025em b}\kern-.08em
    T\kern-.1667em\lower.7ex\hbox{E}\kern-.125emX}}
\begin{document}

\title{Confidence-Aware Self-Distillation for Multimodal Sentiment Analysis with Incomplete Modalities}

\author{Yanxi Luo$^{1,2\ast}$, Shijin Wang$^{1\ast}$, Zhongxing Xu$^{1}$, Yulong Li$^{1 \dag}$, Feilong Tang$^{3 \dag}$, Jionglong Su$^{1 \dag}$ \\
$^1$Xi’an Jiaotong-Liverpool University,
$^2$Renmin University of China,
$^3$Monash University \\}
\maketitle
\renewcommand{\thefootnote}{\fnsymbol{footnote}}
\def\thefootnote{$^{*}$}\footnotetext{Equal contribution. {$\dag$} Corresponding authors.}

\begin{abstract}
Multimodal sentiment analysis (MSA) aims to understand human sentiment through multimodal data. In real-world scenarios, practical factors often lead to uncertain modality missingness. Existing methods for handling modality missingness are based on data reconstruction or common subspace projections. However, these methods neglect the confidence in multimodal combinations and impose constraints on intra-class representation, hindering the capture of modality-specific information and resulting in suboptimal performance. To address these challenges, we propose a Confidence-Aware Self-Distillation (CASD) strategy that effectively incorporates multimodal probabilistic embeddings via a mixture of Student's $t$-distributions, enhancing its robustness by incorporating confidence and accommodating heavy-tailed properties. This strategy estimates joint distributions with uncertainty scores and reduces uncertainty in the student network by consistency distillation. Furthermore, we introduce a reparameterization representation module that facilitates CASD in robust multimodal learning by sampling embeddings from the joint distribution for the prediction module to calculate the task loss. As a result, the directional constraint from the loss minimization is alleviated by the sampled representation. Experimental results on three benchmark datasets demonstrate that our method achieves state-of-the-art performance.
\end{abstract}

\section{Introduction}
Conventional sentiment analysis involves identifying and extracting sentiments or opinions of people from text~\cite{medhat2014sentiment,zhang2025decoding,xue2025mmrc}. Recently, Multimodal Sentiment Analysis (MSA) has garnered increased attention due to the additional information provided by modalities such as facial expressions and voice tone. Previous multimodal research has demonstrated that effectively integrating information from multiple sources into a joint representation leads to more accurate and comprehensive data representations~\cite{wang2024fedmmr,li2024joint,li2025beyond}. Typically, MSA methodologies assume that data from all modalities is available during both the training and inference stages~\cite{li2024unified,li2024correlation,li2025kd}. However, in reality, issues such as noise, data loss, device malfunctions, and privacy concerns can result in incomplete modalities.

\begin{figure}[t]
  \centering
  \begin{tabular}{cc}
    \includegraphics[width=0.47\textwidth]{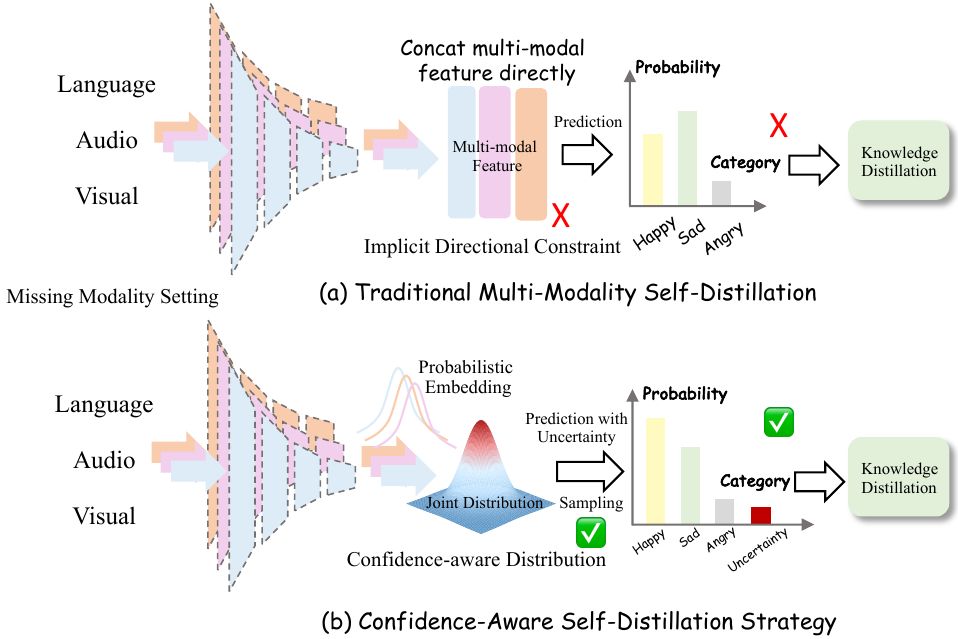} 
  \end{tabular}
  \vspace{-0.1cm}
  \caption{Under the missing-modality setting, traditional methods (a) directly concatenate multimodal features into a single embedding, leading to increased uncertainty and insufficient representation learning, which results in erroneous predictions. In contrast, our CASD (b) integrates confidence-aware distributions by incorporating uncertainty scores from joint distributions into model learning and relaxing intra-class directional constraints, improving the representation ability and producing more reliable predictions.}
  \label{fig1}
  \vspace{-0.5cm}
\end{figure}


To address uncertain missing modalities in MSA, existing approaches fall into two categories: data reconstruction-based methods and common subspace-based methods. Data reconstruction-based methods attempt to reconstruct missing modalities from the available ones~\cite{liu2024modality,wang2024leegnet,xu2024polyp}. While this approach can generate a complete dataset, it often requires substantial computational resources and introduces unwanted noise, which limits its overall effectiveness~\cite{wang2022stepwise}. Meanwhile, common subspace-based methods address the issue by identifying common features that can be shared across all possible combinations of input modalities~\cite{liu2024modality,zhao2024sfc}. These methods aim to project various modalities into a shared latent space. 
However, current methods neglect two key challenges, as shown in Fig.~\ref{fig1} (a): \textbf{(i) Uncertainty in Multimodal Combinations:} Due to the noise in modality data and uncertainty from missing modalities, models that directly concatenate multimodal features can lead to information loss and entanglement. The student model is prevented from learning effective information, resulting in incorrect predictions. 
\textbf{(ii) Implicit Directional Constraint:} Directly projecting different modalities into a deterministic embedding introduces implicit directional constraints. Specifically, samples with different modalities within the same class are forced to learn representations in the same direction. This hinders the model from capturing modality-specific information, resulting in insufficient learning.

In this work, we propose a learning strategy named Confidence-Aware Unified Self-Distillation (CASD) strategy that extracts valuable sentiment information from the confidence distributions of various modalities to alleviate uncertainty caused by missing modalities, as shown in Fig.~\ref{fig1} (b). Specifically, CASD estimates a probabilistic embedding for each modality, rather than a fixed point in the latent space. We employ a Mixture of Student’s $t$-distributions to generate a joint modality distribution that captures unstable heavy-tailed properties using degrees of freedom. These degrees of freedom reflect the confidence of each modality (\textit{i.e.,} higher degrees of freedom indicate higher confidence in modality features), thereby effectively adjusting each modality information. Moreover, we estimate the quality of joint distribution using uncertainty scores from statistical analysis and reduce uncertainty in the student network by consistency distillation.

To address implicit directional constraints and learn diverse features, we introduce a reparameterization representation module (RRM). Specifically, RRM samples embeddings randomly from the joint distribution and inputs them into the prediction module to compute task loss. The sampled representation blocks the directional constraint imposed by loss minimization, preventing the model from relying on a single, fixed embedding direction. Consequently, the model learns embeddings for each modality that capture specific information, enabling the student network to reconstruct valuable missing semantics during distillation.

We evaluate the proposed method under uncertain missing and complete modality conditions in three multimodal benchmarks, achieving state-of-the-art performance. The contributions of this paper are:
1) We introduce a Confidence-Aware Self-Distillation (CASD) strategy that extracts valuable sentiment information from confidence distributions of various modalities to alleviate uncertainty caused by missing modalities and enhance student network robustness.
2) We propose a reparameterization representation module (RRM) that enables CASD to learn robust multimodal joint representations by randomly sampling embeddings from the joint distribution, allowing the model to address implicit directional constraints.
3) Experimental results on benchmark datasets demonstrate that our method significantly improves the efficacy of previous state-of-the-art methods.

\begin{figure*}[t]
    \centering
    \includegraphics[width=0.81\linewidth]{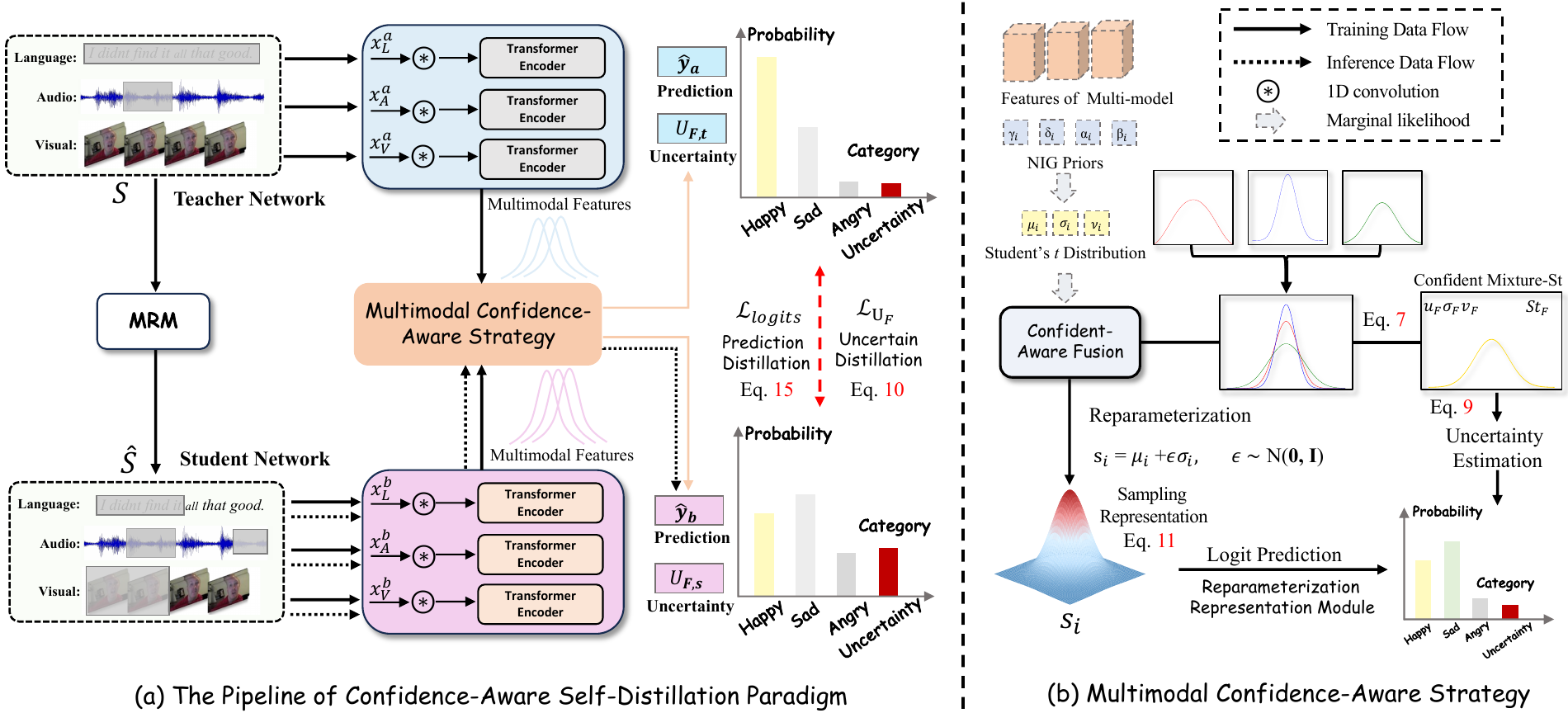}
    \vspace{-0.1cm}
    \caption{Overview of the proposed framework. (a) shows the pipeline of Confidence-Aware Self-Distillation (CASD) paradigm, which regularize student model to reduce the uncertainty of multimodal fusion through an uncertainty consistency loss. (b) illustrates the multimodal confidence-aware strategy, where different colors represent the distributions of different modalities. The reparameterization representation module (RRM) samples embeddings to predict logits and perform distillation.}
    \vspace{-0.4cm}
    \label{fig2}
\end{figure*}

\section{Methodology}
Given a multimodal video dataset $\boldsymbol{D} = \{x_i, y_i\}_{i=1}^N$, where $N$ is the number of samples, each $x_i$ comprises $M$ modality inputs as $x_i = \{x_{i,m}\}_{m=1}^M$, and $y_i=\left\{y_i^c\right\}_{c=1}^C$, where $C$ is the number of categories. 
We define two cases of incomplete modalities to simulate the natural and holistic challenges in real-world scenarios: (i) \textit{intra-modality missingness}, referring to impaired or noisy data within a specific modality, or missing frame-level features. (ii) \textit{inter-modality missingness}, where some modalities are entirely missing. Our goal is to recognize utterance-level sentiments using data with missing modalities.

\subsection{Overall Framework} Fig.~\ref{fig2} (a) illustrates the workflow of the proposed CASD. The teacher network and the student network adopt a consistent structure but have different parameters. During the training phase, our CASD procedure is as follows: (i) we train the teacher network with complete-modality samples and then co-train two models simultaneously. (ii) Given a video segment sample $\boldsymbol{S}$, we generate a missing-modality sample $\hat{\boldsymbol{S}}$ using the Modality Random Missing (MRM) strategy. MRM simultaneously performs intra-modality and inter-modality missing data by replacing the missing features with zero vectors. $\boldsymbol{S}$ and $\hat{\boldsymbol{S}}$ are fed into the initialized student network and the trained teacher network, respectively. (iii) We input the samples $\boldsymbol{S}$ and $\hat{\boldsymbol{S}}$ into the modality representation and construct probabilistic embeddings to achieve a more flexible representation space. (iv) The confidence-aware self-distillation strategy adaptively fuses multimodal probabilistic embeddings, assesses the quality of their joint distributions through uncertainty scores, and ensures consistency by reducing fusion uncertainty in the student network. (v) The RRM samples joint embeddings randomly to predict logits and perform logits distillation, aiming to alleviate the directional constraint on inference representations. During the inference phase, testing samples are only fed into the student network for downstream tasks.

\subsection{Representation Probabilization}
We introduce the extraction and probabilization processes of modality representations using the student network as an example. The incomplete modality $\hat{x}_m$ with $m \in \{L, A, V\}$ is fed into the student network. $\hat{x}_m$ first passes through a 1D temporal convolutional layer with a kernel size $3 \times 3$ to obtain the preliminary representations, denoted as $\boldsymbol{F}_m$. 
We then extend the deep evidential regression model~\cite{amini2020deep} to deep multi-modality evidential classification for MSA. Each $\boldsymbol{F}_m$ is fed into an encoder, which captures the modality dynamics of each sequence through the self-attention mechanism to yield representations $z_m$.

To model the uncertainty for each modality, we build probabilistic embeddings to capture distributions instead of fixed points for each modality, enhancing flexibility by modeling uncertainty. Specifically, we assume the probabilistic embedding $z_m$ follows a multivariate Gaussian distribution:
\begin{equation}\small
p\left(z_m \mid \hat{x}_m\right)=\mathcal{N}\left(z_m \mid \mu_m, \sigma^2_m\right),
\end{equation}
where the mean $\mu_m$ and variance $\sigma_m$ denoting the Gaussian parameters are estimated using the input $\hat{x}_m$. Different from existing methods~\cite{chang2020data,shi2019probabilistic} that estimate these parameters for the feature vector after pooling, we estimate $\mu_m$ and $\sigma_m$ from the feature map directly. These parameters are described by an evidential prior named the Normal-Inverse-Gamma (NIG):
\begin{equation}\small
\operatorname{NIG}\left(\mu_m, \sigma_m^2 \mid \mathbf{p}_m\right)=\mathcal{N}\left(\mu_m \mid \gamma_m, \frac{\sigma_m^2}{\delta_m}\right) \Gamma^{-1}\left(\sigma_m^2 \mid \alpha_m, \beta_m\right),
\end{equation}
where $\Gamma^{-1}$ is an inverse-gamma distribution. Specifically, the multi-evidential heads are placed after the encoders, which outputs the prior NIG parameters $\mathbf{p}_m=\left(\gamma_m, \delta_m, \alpha_m, \beta_m\right)$. Consequently, the Aleatoric Uncertainty (AU) and Epistemic Uncertainty (EU) can be estimated by the mean $\mathbb{E}\left[\sigma_m^2\right]$ and the variance $\operatorname{Var}[\mu_m]$, respectively:
\begin{equation}\small
\mathrm{AU}=\mathbb{E}\left[\sigma_m^2\right]=\frac{\beta_m}{\alpha_m-1}, \quad \mathrm{EU}=\operatorname{Var}[\mu_m]=\frac{\beta_m}{\delta_m\left(\alpha_m-1\right)}.
\end{equation}
where $\mathbb{E}[\sigma_m^2]$ captures inherent data randomness, making it suitable for AU, while $\operatorname{Var}[\mu_m]$ quantifies the spread of $\mu_m$, reflecting model confidence and thus representing EU. This decomposition ensures interpretable and robust uncertainty estimation.
Subsequently, the Student's $t$ predictive distributions are derived from the interaction of the prior and the Gaussian likelihood of each modality, given by:
\begin{equation}\small
p\left(z_m \mid \mathbf{p}_m\right) =\frac{p\left(z_m \mid \theta, \mathbf{p}_m\right) p\left(\theta \mid, \mathbf{p}_m\right)}{p\left(\theta \mid z_m, \mathbf{p}_m\right)}.
\end{equation}
When an NIG prior is applied to our Gaussian likelihood function, the resulting analytical solution for the Student's $t$ predictive distributions is:
\begin{equation}\small
p\left(z_i^m \mid \mathbf{p}_m\right) =\operatorname{St}\left(z_i^m ; \gamma_m, o_m, 2 \alpha_m\right),
\end{equation}
where $o_m=\frac{\beta_m\left(1+\delta_m\right)}{\delta_m \alpha_m}$. Thus, the distributions of the three modalities are transformed into the Student's $t$ distributions $S \mathrm{t}\left(z_m ; u_m, \sigma_m, v_m\right)=S \mathrm{t}\left(z_m ; \gamma_m, \frac{\beta_m\left(1+\delta_m\right)}{\delta_m \alpha_m}, 2 \alpha_m\right)$. 

\subsection{Confidence-Aware Self-Distillation Strategy}
We focus on integrating multiple $\mathrm{S}t$ distributions from different modalities into a unified $\mathrm{S}t$. To this end, the joint modality of distribution can be denoted as:
\begin{equation}\small
p\left(x_1, x_2, x_3\right)=S \mathrm{t}\left(z_m ; u_F, \Sigma_F, v_F\right),
\end{equation}
where
$
u_F = \begin{bmatrix}\small
u_1 \\
u_2 \\
u_3
\end{bmatrix},
\quad
\Sigma_F = \begin{bmatrix}\small
\Sigma_1 \\
\Sigma_2 \\
\Sigma_3
\end{bmatrix},
\quad
v_F = \begin{bmatrix}\small
v_1 \\
v_2 \\
v_3
\end{bmatrix}.
$
To preserve the closed form of the Student's $t$ distribution and maintain the heavy-tailed properties of the fused modality, the updated parameters are described by~\cite{roth2013student}. Specifically, we adjust the degrees of freedom of the distributions to ensure consistency. As described in~\cite{roth2013student}, smaller degrees of freedom correspond to heavier tails, while larger degrees of freedom indicate lighter tails but better overall tail behavior. Furthermore, the variance of the Student's $t$ distribution decreases as the degrees of freedom $v$ increase, which reflects higher confidence. We assume that the fused Student's $t$ distribution remains approximately a Student's $t$ distribution. Assuming that the degrees of freedom of $v_1$, $v_2$ and $v_3$ are adjusted such that the resulting fused Student's $t$ distribution, $S \mathrm{t}\left(z_m ; u_F, \Sigma_F, v_F\right)$ will be updated as:
\begin{equation}\small
\left\{\begin{array}{c}
v_F = \text{min}(v_1, v_2, v_3) \\
u_F = C_1 u_1 + C_2 u_2 + C_3 u_3 \\
\Sigma_F = \frac{1}{3}\left(\Sigma_1 + \frac{v_2\left(v_1 - 2\right)}{v_1\left(v_2 - 2\right)} \Sigma_2 + \frac{v_3\left(v_1 - 2\right)}{v_1\left(v_3 - 2\right)} \Sigma_3\right)
\end{array},\right.
\end{equation}
where $C_1$, $C_2$ and $C_3$ denote the confidence from the distribution of uni-modality, which can be defined as:
\begin{equation}\small
\mathcal{C}_1=\frac{v_1}{v_1+v_2}, \quad \mathcal{C}_2=\frac{v_2}{v_1+v_2} \quad \mathcal{C}_3 = \frac{v_3}{v_1 + v_2 + v_3}.
\end{equation}
Therefore, the uncertainty score $U_F$ for the fused modality can be estimated by:
\begin{equation}\small
\begin{aligned}
u_F = \int z_m p\left(z_m \mid x_F, \mathbf{p}_F\right) dz_m \\
U_F=\Sigma_F \frac{v_F}{v_F-3}=\Sigma_F\left(1+\frac{3}{v_F-3}\right)
\end{aligned},\label{10}
\end{equation}
where $x_F$ denotes the fused modality data combines multiple modalities into one representation. $u_F$ is the mean of the fused modality’s Student's $t$ distribution. $\mathbf{p}_F=$ $\left(u_F, \sigma_F, v_F\right)$ is the parameter of the St distribution after fusion. The term $\frac{v_F}{v_F-3}$ adjusts for the degrees of freedom, reflecting the heavy-tailed nature of the Student's $t$ distribution. Confidence-aware fusion can be seen in Fig.~\ref{fig1} (b).

To distill the uncertainty $U_F$, we can minimize the uncertainty difference between the teacher network and the student network using the Mean Squared Error. Specifically, the goal is to make the student's uncertainty estimate as close as possible to that of the teacher.
\begin{equation}\small
\mathcal{L}_{U_F}=\frac{1}{N} \sum_{i=1}^N\left\|U_{F, s}^{(i)}-U_{F, t}^{(i)}\right\|^2,
\label{11}
\end{equation}
where $\|\cdot\|^2$ represents $\ell_2$ norm function. $U_{F,s}^{(i)}$ and $U_{F,t}^{(i)}$ represent the estimates of uncertainty for the $i$-th sample by the teacher and student networks, respectively.

\subsection{Reparameterization Representation Module}

Traditional common subspace methods limit intra-class representation limits, which forces samples with different modalities within the same class to learn similar representations. This limitation reduces the capacity to capture modality-specific characteristics, thereby impairing the diversity and effectiveness of multimodal feature integration. To address this, we propose a reparameterization representation module that alleviates these constraints by sampling modality-specific representations. Since directly sampling from these distributions is non-differentiable, it becomes a challenge in the training process of the model. Therefore, we introduce the reparameterization trick~\cite{li2018smoothing}, which makes the sampling operation differentiable, allowing for the effective optimization of model parameters during backpropagation. We employ the fused distribution $\mathbf{p}_F$ and implement the sampling operation through the reparameterization trick as:
\begin{equation}\small
s=u_F+\sigma_F \cdot t, \quad where \quad t \sim \operatorname{St}\left(v_F\right),
\label{12}
\end{equation}
where $t$ is sampled from the standard Student's $t$-distribution $\operatorname{St}(0, 1, v_F)$, and the embedding $s$ is generated using Eq.~\ref{12}, instead of directly sampling from $\operatorname{St}\left(u_F, \sigma_F^2, v_F\right)$. During the training phase, the sampled representation $s_i$ is used to train the model, allowing it to adapt to the variability and uncertainty across different modality combinations, thereby learning more robust representations. In the inference phase, we use the fused mean $u_F$ as the final representation $\mu_i$, ensuring stability and accuracy.
Therefore, the typical cross-entropy loss for the model is:
\begin{equation}\small
\mathcal{L}_{\text{CE}}=-\frac{1}{N} \sum_{i=1}^N \log \frac{e^{W_{y} \cdot g\left(s\right)}}{\sum_{k=1}^M e^{\left(W_k \cdot g\left(s\right)\right)}},
\end{equation}
where $W$ denotes the parameter matrix of the final linear classifier, and $g\left(s\right)$ represents the feature vector $s$ after global average pooling and flattening. The logit $\hat{y}$, defined as $W_{y} \cdot g\left(s\right)$, represents the raw score for the true class $y$ before applying softmax.

\noindent \textbf{Analysis:} This only requires the sampled embedding $g\left(s\right)$ to share the same direction with $W_{y}$. The inference embedding $\mu_F$ for different input combinations belonging to the same class could be non-parallel. This relaxes the directional constraint on the inference representation and enables the model to capture the specific information for different modality combinations. In particular, the value of $\sigma_F$ controls the degree of relaxation. When $\sigma_F=0, s$ will equal $\mu_F$, which degenerates into the vanilla subspace-based methods without relaxation. In contrast, a larger $\sigma_F$ increases the sampling range, making the directional constraint from $s$ to $\mu_F$ weaker.

\subsection{Self-Distillation Optimization Paradigm}
To align the distributions of the logits from the teacher and student models, we use a logits distillation mechanism that effectively transfers knowledge. This mechanism is versatile and can be applied to classification tasks. We employ the Jensen-Shannon (JS) divergence as a measure of discrepancy between the teacher and student logits distributions. This divergence overcomes the asymmetry issues of Kullback-Leibler (KL) divergence and provides a more balanced measure of distributional differences. The KL divergence is defined as:

\begin{equation}\small
\mathcal{D}_{K L}\left(\boldsymbol{p}_b \| \boldsymbol{p}_a\right)=-\frac{1}{n} \sum_{i=1}^n \boldsymbol{p}\left(\boldsymbol{x}_b\right) \log \frac{\boldsymbol{p}\left(\boldsymbol{x}_a\right)}{\boldsymbol{p}\left(\boldsymbol{x}_b\right)},
\end{equation}
where $\boldsymbol{p}_b$ is the target probabilities as soft labels to supervise the learning of the predicted probabilities $\boldsymbol{p}_a$. The logits distillation loss is denoted as:
\begin{equation}\small
\mathcal{L}_{\text {logits }} =\mathcal{D}_{JS}\left(\hat{y}_a \| \hat{y}_b\right) =\frac{1}{2}\left(\mathcal{D}_{K L}\left(\hat{y}_a \| \boldsymbol{M}\right)\right)+\mathcal{D}_{K L}\left(\hat{y}_b \| \boldsymbol{M}\right)),
\end{equation}
where $\hat{y}_a$ and $\hat{y}_b$ are the logits from the student and teacher models, respectively. $\boldsymbol{M}$ is the average distribution of $\hat{y}_a$ and $\hat{y}_b$. The overall training objective $\mathcal{L}_{\text{total}}$ is expressed as:
\begin{equation}\small
\mathcal{L}_{\text{total}}=\mathcal{L}_{\text{CE}}+\alpha \mathcal{L}_{\text{logits}}  +\beta \mathcal{L}_{U_F},
\end{equation}
where $\mathcal{L}_{\mathrm{CE}}$ is the cross-entropy loss, $\mathcal{L}_{U_F}$ is the uncertainty distillation loss, and $\mathcal{L}_{\text {logits }}$ is the logits distillation loss with $\alpha$ and $\beta$ as their respective weights.

\section{Experiments}

\subsection{Datasets and Evaluation Metrics}
We conduct experiments on three benchmark MSA datasets: MOSI~\cite{zadeh2016mosi}, MOSEI~\cite{zadeh2018multimodal}, and IEMOCAP~\cite{busso2008iemocap}. The MOSI dataset comprises 2,199 video clips capturing authentic opinions, with 1,284 clips allocated for training, 229 for validation, and 686 for testing. MOSI and MOSEI involve video clips labeled with sentiment scores ranging from -3 to +3, with performance assessed using Mean Absolute Error (MAE) and F1 score for binary sentiment classification. The IEMOCAP dataset, which consists of conversational videos, is utilized for emotion recognition tasks. Following~\cite{wang2019words}, four emotional categories are classified: happiness, sadness, anger, and neutrality. Model performance on the IEMOCAP dataset is assessed using the F1 score as the primary metric.

\begin{figure}[t]
    \centering
    \includegraphics[width=0.37\textwidth]{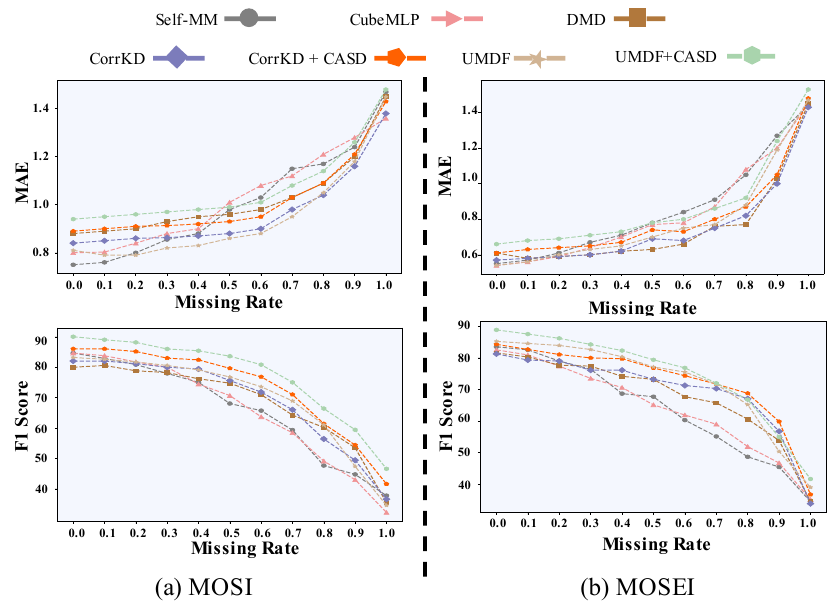}
    \vspace{-0.3cm}
    \caption{Comparison of MAE and F1 scores at various missing rates on (a) MOSI and (b) MOSEI.}
    \vspace{-0.45cm}
    \label{fig:IEMOCAP}
\end{figure}

\subsection{Comparison with State-of-the-Art Methods}

We compare CASD with seven representative SOTA methods, including complete-modality methods: Self-MM~\cite{yu2021learning}, CubeMLP~\cite{sun2022cubemlp}, and DMD~\cite{li2023decoupled}, and missing-modality methods: 1) joint learning methods (\textit{i.e.,} UMDF~\cite{li2024unified}, MCIS-MMIM~\cite{yang2024towards} and CorrKD~\cite{li2024correlation}), and 2) generative methods (\textit{i.e.,} SMIL~\cite{ma2021smil} and GCNet~\cite{lian2023gcnet}). Extensive experiments evaluate the robustness and effectiveness of CASD in the cases of intra-modality and inter-modality missingness.

\noindent\textbf{Intra-modality missing robustness.} We randomly drop frame-level features in modality sequences with a drop ratio $p \in \{0.1, 0.2, \cdots, 1.0$\} to simulate testing conditions of intra-modality missingness. Fig.~\ref{fig:IEMOCAP} shows the performance curves for different $p$ values, reflecting model robustness. Key observations include:
(i) Increasing $p$ reduces performance across all models, highlighting the impact of intra-modality missingness on sentiment semantics and joint multimodal representations.
(ii) Compared with complete-modality methods (\textit{i.e.,} Self-MM, CubeMLP, and DMD), CASD excels under missing-modality conditions and remains competitive with complete modalities, as it effectively captures and reconstructs sentiment semantics from incomplete data. 
(iii) Compared to other missing-modality methods, CASD demonstrates superior robustness by leveraging confidence-aware self-distillation to integrate multimodal probabilistic embeddings and handle uncertainty, enabling the student network to reconstruct missing semantics and generate robust representations.

\begin{table}[t]
    \centering
    \tiny
    \caption{Performance comparison of different models on MOSI and MOSEI datasets under various testing conditions.}
    \vspace{-0.2cm}
    \resizebox{0.48\textwidth}{!}{
    \begin{tabular}{clcccccccccccccccc}
    \toprule
     & \multirow{2}{*}{Models} & \multicolumn{8}{c}{Testing Conditions} \\ \cmidrule(r){3-10}
     & & \textit{\{l\}} & \textit{\{a\}} & \textit{\{v\}} & \textit{\{l,a\}} & \textit{\{l,v\}} & \textit{\{a,v\}} & \textbf{\textit{Avg.}} & \textit{\{l,a,v\}} \\ \midrule
    \multirow{9}{*}{\rotatebox{90}{MOSI}}
     & Self-MM & 67.80 & 40.95 & 38.52 & 69.81 & 74.97 & 47.12 & 56.53 & 84.64 \\
     & CubeMLP & 64.15 & 38.91 & 43.24 & 63.76 & 65.12 & 47.92 & 53.85 & 84.57 \\
     & DMD & 68.97 & 43.33 & 42.26 & 70.51 & 68.45 & 50.47 & 57.33 & 84.50 \\
     & GCNet & 80.91 & 65.07 & 58.70 & 84.73 & 83.58 & 70.02 & 73.84 & 83.20 \\
     & SMIL & 78.26 & 67.69 & 59.67 & 79.82 & 79.15 & 71.24 & 72.64 & 82.85 \\
     & MCIS-MMIM & - & - & - & - & - & - & - & 86.5  \\ \cmidrule(r){2-10}
     & CorrKD & 81.20 & 66.52 & 60.72 & 83.56 & 82.41 & 73.74 & 74.69 & 83.94 \\
     & \cellcolor{gray!30} + CASD  & \cellcolor{gray!30} \textbf{82.11} & \cellcolor{gray!30} \textbf{68.51} & \cellcolor{gray!30} \textbf{62.59} & \cellcolor{gray!30} \textbf{85.21} & \cellcolor{gray!30} \textbf{84.52} & \cellcolor{gray!30} \textbf{74.88} & \cellcolor{gray!30} \textbf{76.29} & \cellcolor{gray!30} \textbf{86.03} \\
     & UMDF & 82.92 & 67.80 & 59.92 & 85.63 & 84.09 & 72.98 & 75.56 & 83.36 \\ 
     & \cellcolor{gray!30} + CASD  & \cellcolor{gray!30} \textbf{84.23} & \cellcolor{gray!30} \textbf{69.92} & \cellcolor{gray!30} \textbf{62.48} & \cellcolor{gray!30}\textbf{87.01} & \cellcolor{gray!30}\textbf{86.07} & \cellcolor{gray!30}\textbf{74.31} & \cellcolor{gray!30}\textbf{77.63} & \cellcolor{gray!30}\textbf{85.95} \\
     \midrule
    \multirow{9}{*}{\rotatebox{90}{MOSEI}} 
     & Self-MM & 71.53 & 43.57 & 37.61 & 75.91 & 74.62 & 49.52 & 58.79 & 83.69 \\
     & CubeMLP & 67.52 & 39.54 & 32.58 & 71.69 & 70.06 & 48.54 & 54.99 & 83.17 \\
     & DMD & 70.26 & 46.18 & 39.84 & 74.78 & 72.45 & 52.70 & 59.37 & 84.78\\
     & GCNet & 80.52 & 66.54 & 61.83 & 81.96 & 81.15 & 69.21 & 73.54 & 82.35 \\
     & MCIS-DMD & - & - & - & - & - & - & - & 87.1  \\ \cmidrule(r){2-10}
     & CorrKD & 80.76  & 66.09  & 62.30  & 81.74  & 81.28  & 71.92  & 74.02  & 82.16 \\ 
     & \cellcolor{gray!30} CorrKD+CASD & \cellcolor{gray!30} \textbf{81.85} & \cellcolor{gray!30} \textbf{67.96} & \cellcolor{gray!30} \textbf{63.08} & \cellcolor{gray!30} \textbf{83.28} & \cellcolor{gray!30} \textbf{82.69} & \cellcolor{gray!30} \textbf{72.84} & \cellcolor{gray!30} \textbf{75.23} & \cellcolor{gray!30} \textbf{84.89} \\
     & UMDF & 81.57 & 67.42 & 61.57 & 83.25 & 82.14 & 69.48 & 74.24 & 82.16 \\ 
     & \cellcolor{gray!30} UMDF+CASD & \cellcolor{gray!30} \textbf{83.43} & \cellcolor{gray!30} \textbf{69.94} & \cellcolor{gray!30} \textbf{64.27} & \cellcolor{gray!30} \textbf{85.61} & \cellcolor{gray!30} \textbf{84.51} & \cellcolor{gray!30} \textbf{73.98} & \cellcolor{gray!30} \textbf{76.93} & \cellcolor{gray!30} \textbf{85.93} \\ \midrule
\end{tabular}}
\vspace{-0.1cm}
\label{tab:label2}
\end{table}

\noindent \textbf{Inter-modality missing robustness.} Tables~\ref{tab:label2} and ~\ref{tab:label1} simulate missing modality conditions. “${l}$” indicates only the language modality is available, “${l, a, v}$” represents full modality availability, and “Avg.” reflects average performance across six missing modality scenarios. Key insights:
(i) Missing modalities reduce model performance, highlighting the importance of integrating diverse modal information for enhancing emotional semantics.
(ii) Under missing modality conditions, CASD+CorrKD, integrated as a plugin, outperforms in most metrics, demonstrating robustness. On the MOSI dataset, it improves the average F1 score by 2.45\% over GCNet and 4.86\% when the language modality is missing (${a, v}$), benefiting from CASD's confidence-aware strategy for capturing and integrating multimodal features.
(iii) CASD acts as a plug-and-play module, boosting model performance under missing modality conditions. On the MOSEI dataset, integrating CASD with UMDF improves the F1 score by 3.77\% under full modalities and 4.5\% when the language modality is missing, demonstrating its effectiveness in enhancing robustness and optimizing semantic inference.

\begin{table}[t]
\centering
\small
\caption{Performance comparison under different testing conditions of intermodality missingness on IEMOCAP.}
\vspace{-0.1cm}
\resizebox{0.48\textwidth}{!}{
\begin{tabular}{lccccccccc}
\midrule
\multirow{2}{*}{Models} & \multirow{2}{*}{Metrics} & \multicolumn{8}{c}{Testing Conditions} \\ \cmidrule(r){3-10}
 & & \makebox[0pt][c]\{\textit{ l }\} & \{\textit{ a }\} & \{\textit{ v }\} & \{\textit{ l,a }\} & \{\textit{ l,v }\} & \{\textit{ a,v }\} & \textit{Avg.} & \{\textit{ l,a,v }\} \\ \midrule

\multirow{4}{*}{CubeMLP} & Happy & 68.9 & 54.3 & 51.4 & 72.1 & 69.8 & 60.6 & 89.0 & 62.9 \\
 & Sad & 65.3 & 54.8 & 53.2 & 70.3 & 68.7 & 58.1 & \textbf{88.5} & 61.7 \\
 & Angry & 65.8 & 53.1 & 50.4 & 69.5 & 69.0 & 54.8 & 87.2 & 61.8 \\
 & Neutral & 53.5 & 50.8 & 48.7 & 57.3 & 54.5 & 51.8 & 71.8 & 52.8 \\ \midrule

\multirow{4}{*}{GCNet} & Happy & 81.9 & 67.3 & 66.6 & 83.7 & 82.5 & 69.8 & 87.7 & 75.3 \\
 & Sad & 80.5 & 69.4 & 66.1 & 83.8 & 82.1 & 70.5 & 86.9 & 75.4 \\
 & Angry & 80.1 & 66.2 & 64.2 & 82.5 & 81.6 & 68.1 & 85.2 & 73.8 \\
 & Neutral & 61.8 & 51.1 & 49.6 & 63.5 & 53.3 & 53.3 & 71.1 & 57.6 \\ \midrule
\multirow{4}{*}{UMDF} & Happy & 82.4 & 68.6 & 67.2 & 85.9 & 84.2 & 69.1 & 87.9 & 76.2 \\
 & Sad & 81.2 & 70.7 & 67.1 & 83.6 & 82.2 & 71.9 & 86.5 & 76.1 \\
 & Angry & 81.6 & 67.9 & 65.1 & 83.9 & 82.5 & 67.9 & 85.8 & 74.6 \\
 & Neutral & 64.3 & 53.2 & 50.9 & 67.2 & 65.3 & 55.0 & 70.5 & 59.3 \\ \midrule 
\multirow{4}{*}{UMDF+CASD} & \cellcolor{gray!30} Happy & \cellcolor{gray!30}84.3 & \cellcolor{gray!30}71.6 & \cellcolor{gray!30}70.1 & \cellcolor{gray!30}87.2 & \cellcolor{gray!30}86.3 & \cellcolor{gray!30}72.9 & \cellcolor{gray!30}89.6 & \cellcolor{gray!30}78.7 \\
& \cellcolor{gray!30}Sad & \cellcolor{gray!30}83.7 & \cellcolor{gray!30}74.0 & \cellcolor{gray!30}69.8 & \cellcolor{gray!30}85.8 & \cellcolor{gray!30}84.9 & \cellcolor{gray!30}74.7 & \cellcolor{gray!30}88.1 & \cellcolor{gray!30}77.6 \\
& \cellcolor{gray!30}Angry & \cellcolor{gray!30}83.9 & \cellcolor{gray!30}70.5 & \cellcolor{gray!30}68.1 & \cellcolor{gray!30}86.6 & \cellcolor{gray!30}85.0 & \cellcolor{gray!30}72.3 & \cellcolor{gray!30}87.4 & \cellcolor{gray!30}76.4 \\
& \cellcolor{gray!30}Neutral & \cellcolor{gray!30}65.4 & \cellcolor{gray!30}56.4 & \cellcolor{gray!30}54.4 & \cellcolor{gray!30}70.6 & \cellcolor{gray!30}66.5 & \cellcolor{gray!30}59.2 & \cellcolor{gray!30}62.8 & \cellcolor{gray!30}73.5 \\
\midrule
\end{tabular}}
\vspace{-0.5cm}
\label{tab:label1}
\end{table}

\subsection{Ablation Studies}
\noindent \textbf{Effectiveness of each component.} Table~\ref{component} presents an ablation study to evaluate our method. We use the full modality method (Experiment I) as the baseline for comparison. UMDF serves as the baseline method for modality missingness (Experiment II). In Experiment III, we introduce the Confidence-Aware Strategy, which provides confidence scores for each modality to generate the joint distribution, resulting in a significant performance improvement of +1.54\% under six different missing modality testing conditions. In Experiment IV, by incorporating the joint distribution uncertainty loss $\mathcal{L}_{U_F}$, performance improves further by +0.65\% under missing modality and +1.13\% under complete modality, allowing the student network to better adapt to various modality missing scenarios and enhancing its generalization ability to unknown data. Experiment V introduces the PPM, adding a +0.5\% improvement by sampling joint distributions, which alleviates representation constraints and improves emotional semantic reconstruction.

\begin{table}[t]
  \begin{minipage}[p]{0.5\textwidth}
    \centering
    \tiny
    \caption{Ablation study on main components of the proposed framework on testing conditions of inter-modality missingness on MOSEI. UMDF: Plain distillation learning. Confidence-Aware: Provides confidence scores for each modality and elegantly integrates multimodal. $\mathcal{L}_{U_F}$: Estimate joint distributions with uncertainty scores and distill. PPM: Randomly sampling the joint distribution to alleviate the representation constraints.} 
    \vspace{-0.2cm}
    \resizebox{\textwidth}{!}{
    \begin{tabular}{c|c|c|c|c|c|ccc|ccc}
    \toprule[1pt]
    & UMDF  & Confidence-Aware & $\mathcal{L}_{U_F}$ & PPM & Avg. & \{l, a, v\} \\
    \midrule I & &  &  & &  72.12 & 80.45   \\
    II & \Checkmark &  &  & &  74.24 & 82.16   \\
    III & \Checkmark & \Checkmark & & &  75.78 & 83.54   \\
    IV & \Checkmark & \Checkmark & \Checkmark & &  76.43 & 84.67   \\
    V & \Checkmark & \Checkmark & \Checkmark & \Checkmark &  76.93 & 85.93   \\
    \toprule[1pt]
    \end{tabular}
    \vspace{-0.3cm}}
    \label{component}
  \end{minipage}
\end{table}
\begin{table}[t]
\centering
\tiny
\vspace{-0.2cm}
\caption{Performance comparison of methods for estimating the mean and variance of the probabilistic distribution.}
\vspace{-0.1cm}
\resizebox{0.48\textwidth}{!}{
\begin{tabular}{lcccccccc}
\midrule
\multirow{2}{*}{Models} & \multicolumn{7}{c}{\textbf{Testing Conditions}} & \\ \cmidrule(lr){2-9}
 & \{l\} & \{a\} & \{v\} & \{l, a\} & \{l, v\} & \{a, v\} & Avg. & \{l, a, v\} \\ \midrule
UMDF & 81.57 & 67.42 & 61.57 & 83.25 & 82.14 & 69.48 & 74.24 & 82.16 \\
+PE & 82.10 & 68.55 & 62.30 & 84.10 & 83.00 & 70.50 & 75.20 & 83.00 \\
+PCME & 82.85 & 69.12 & 63.05 & 84.80 & 83.75 & 71.25 & 75.40 & 83.55 \\
\rowcolor{gray!30} + CASD & \textbf{83.43} & \textbf{69.94} & \textbf{64.27} & \textbf{85.61} & \textbf{84.51} & \textbf{73.98} & \textbf{76.93} & \textbf{85.93} \\ \midrule
\end{tabular}}
\label{table:distribution}
\vspace{-0.4cm}
\end{table}

\noindent \textbf{Comparison of different distribution estimation.} We compare distribution estimation methods under inter-modality missingness on the MOSEI dataset, as shown in Table~\ref{table:distribution}. “PE” denotes the traditional method that estimates the distribution of feature vectors using a fully connected layer~\cite{shi2019probabilistic}. “PCME” introduces attention modules to aggregate information from the feature map to estimate the distribution of feature vectors~\cite{chun2021probabilistic}. CASD estimates the distribution of the feature map directly using Student’s $t$ distributions. As shown in Table~\ref{table:distribution}, the CASD method improves upon “PE” and “PCME” under missing modalities and under complete modalities. This highlights CASD's effectiveness in capturing more detailed distribution information.

\vspace{-0.2cm}
\section{Conclusions}
In this paper, we propose a novel Confidence-Aware Self-Distillation (CASD) strategy for MSA, addressing challenges of insufficient representation learning and implicit directional constraints in directly concatenated and projected multimodal features. Our method effectively integrates multimodal probabilistic embeddings and estimates joint distributions with uncertainty scores, reducing uncertainty in the student network through consistency distillation. The RRM enhances learning by sampling embeddings from the joint distribution, overcoming implicit directional constraints. Experiments on three benchmarks demonstrate state-of-the-art performance, with ablation studies validating its effectiveness.

\vspace{-0.2cm}
\bibliographystyle{IEEEbib}
\bibliography{icme2025references}

\end{document}